\PassOptionsToPackage{hyphens}{url}
\documentclass[11pt]{article}

\usepackage[margin=1in]{geometry}
\usepackage[T1]{fontenc}
\usepackage[utf8]{inputenc}
\usepackage{lmodern}
\usepackage{microtype}
\usepackage{graphicx}
\usepackage{float}
\usepackage{booktabs}
\usepackage{tabularx}
\usepackage{longtable}
\usepackage{array}
\usepackage{ragged2e}
\usepackage{authblk}
\usepackage{hyperref}
\usepackage[hyphens]{url}
\usepackage{enumitem}
\usepackage{amsmath}
\hypersetup{colorlinks=true, linkcolor=blue, urlcolor=blue, citecolor=blue}
\setlist[itemize]{noitemsep, topsep=2pt}
\setlist[enumerate]{noitemsep, topsep=2pt}
\newcolumntype{Y}{>{\RaggedRight\arraybackslash}X}
\newcolumntype{C}[1]{>{\Centering\arraybackslash}p{#1}}
\title{Exploring Data Augmentation and Resampling Strategies for Transformer-Based Models to Address Class Imbalance in AI Scoring of Scientific Explanations in NGSS Classroom}
\date{}
\author[1]{Prudence Djagba\thanks{\texttt{djagbapr@msu.edu}}}
\author[1]{Kevin Haudek\thanks{\texttt{haudekke@msu.edu}}}
\author[2]{Clare G.C. Franovic\thanks{\texttt{carls500@msu.edu}}}
\author[3]{Leonora Kaldaras\thanks{\texttt{lkaldara@Central.UH.EDU}}}
\affil[1]{CREATE for STEM, Michigan State University, East Lansing, USA}
\affil[2]{Human Biology, Michigan State University, East Lansing, USA}
\affil[3]{University of Houston, Houston, USA}
\begin{document}
\maketitle
\textbf{This work is presented and published as a conference paper at NARST, April 19–22, 2026,  held at Sheraton Grand Seattle 1400 Sixth Ave, Seattle, WA 98101, USA.}
\begin{abstract}
Automated scoring of students' scientific explanations offers the potential for immediate, accurate feedback, yet class imbalance in rubric categories particularly those capturing advanced reasoning remains a challenge. This study investigates augmentation strategies to improve transformer-based text classification of student responses to a physical science assessment based on an NGSS-aligned learning progression. The dataset consists of 1,466 high school responses scored on 11 binary-coded analytic categories. This rubric identifies six important components including scientific ideas needed for a complete explanation along with five common incomplete or inaccurate ideas. Using SciBERT as a baseline, we applied fine-tuning and test these augmentation strategies: (1) GPT-4--generated synthetic responses, (2) EASE, a word-level extraction and filtering approach, and (3) ALP (Augmentation using Lexicalized Probabilistic context-free grammar) phrase-level extraction.

While fine-tuning SciBERT improved recall over baseline, augmentation substantially enhanced performance, with GPT data boosting both precision and recall, and ALP achieving perfect precision, recall, and F1 scores across most severe imbalanced categories (5,6,7 and 9). Across all rubric categories EASE augmentation substantially increased alignment with human scoring for both scientific ideas (Categories 1--6) and inaccurate ideas (Categories 7--11). We compared different augmentation strategies to a traditional oversampling method (SMOTE) in an effort to avoid overfitting and retain novice-level data critical for learning progression alignment. Findings demonstrate that targeted augmentation can address severe imbalance while preserving conceptual coverage, offering a scalable solution for automated learning progression-aligned scoring in science education.
\end{abstract}

\section{Introduction}
Data augmentation has become an important strategy for improving the performance of machine learning models by increasing the effective size of training datasets without relying on additional external data sources [13]. In this context, data augmentation refers to systematically expanding the original training data through transformations of existing examples, allowing models to learn more robust and generalizable patterns. This approach has been shown to be especially beneficial for deep learning models, which typically perform better when trained on larger and more diverse datasets. In text classification research, where labeled data are often limited or unevenly distributed across categories, data augmentation has gained growing attention. Prior work has demonstrated that augmentation techniques can improve model performance in educational settings, including student achievement classification [14], prediction of at-risk academic outcomes [15], and cheating detection [16], underscoring the relevance of augmentation methods for education-focused machine learning studies.

More recently, large language models (LLMs) such as BERT, XLNet, LaMDA, LLaMA, and GPT have introduced new possibilities for data augmentation due to their advanced natural language generation capabilities. Among these models, BERT has been widely adopted for representation learning, while generative models such as GPT-4 and ChatGPT have further expanded augmentation strategies. Released by OpenAI and trained using reinforcement learning from human feedback, ChatGPT and GPT-4 produce more accurate and contextually coherent text than earlier LLMs, making them particularly well suited for generating high-quality synthetic data [18]. Consequently, researchers have increasingly leveraged GPT-based models for data augmentation in a range of natural language processing tasks [1,3].

The advancement of automated text classification has opened new opportunities in educational assessment, particularly in the context of evaluating scientific reasoning of student responses in science education. We are using learning progressions (LPs) as a guiding framework for structuring science assessments [30]. LPs describe increasingly sophisticated ways of thinking about core scientific ideas, allowing educators to better interpret student understanding over time. As such, LPs are grounded in a developmental approach to learning, to which assessments can be aligned to gauge student learning [12]. When scoring is aligned with LP levels, assessments not only measure performance but also guide instruction by highlighting where students are on a conceptual continuum [8]. A persistent issue in automated scoring is the class imbalance across rubric categories, where critical dimensions of reasoning are underrepresented in the training data. This imbalance leads to biased models that tend to favor majority classes [3]. This challenge is particularly pronounced in LP-aligned assessments, where we do not expect all students to hold similar ideas at the same time, and the most sophisticated ideas in higher LP levels may occur relatively infrequently during early stages of learning [29]. As a result, addressing class imbalance is essential to ensure that automated scoring systems can accurately capture the full range of reasoning envisioned by LP frameworks.

To address these challenges, we explore how transformer-based models---known for their state-of-the-art performance in natural language processing (NLP) can be leveraged for automated scoring of student explanations in science education [5]. Recent studies have demonstrated that transformer architectures such as BERT and its domain-specific variants can effectively capture semantic and contextual features in student-generated text, leading to substantial improvements over traditional feature-based or shallow learning approaches in automated scoring tasks [22, 23]. In science assessment contexts, transformers have been successfully applied to score constructed responses aligned with analytic rubrics, showing strong agreement with human raters and improved sensitivity to conceptual reasoning rather than surface-level language features [24, 25]. These models are particularly well suited for science explanations, which often involve complex causal structures, discipline-specific vocabulary, and implicit reasoning patterns.

Further, this alignment between scores, assessment target and LP requires analytic rubrics capable of capturing specific features of student thinking features that are often embedded in rich, varied natural language explanations. This demand is particularly evident in the context of three-dimensional (3D) science assessments, which integrate disciplinary core ideas (DCIs), science and engineering practices (SEPs), and crosscutting concepts (CCCs) as called for by the Next Generation Science Standards (NGSS). Scoring such assessments is inherently complex, requiring careful analysis of student responses that reflect multiple dimensions of reasoning [27, 29, 30]. Human scoring of these assessments is labor-intensive and susceptible to inconsistency, while automated scoring faces technical and validity challenges such as nuanced language variation, limited labeled data, and class imbalance, especially in categories targeting advanced scientific thinking [28].

However, despite advances in automated scoring using NLP, prior work also highlights persistent limitations. Transformer-based models often struggle with severely imbalanced rubric categories, especially those representing advanced or less frequently expressed forms of scientific reasoning [21, 26]. As a result, models achieved high overall accuracy while performing poorly on critical learning progression levels, limiting their usefulness for formative feedback [30]. Additionally, many existing studies rely on relatively large datasets or focus on single scoring dimensions, leaving open questions about scalability and robustness in small educational datasets. These challenges motivate the need for technical strategies such as fine-tuning and targeted data augmentation that improve predictive model outcomes, especially on underrepresented but instructionally meaningful categories.

To address these challenges, we explore how transformer-based models can be fine-tuned to classify student explanations into analytic rubric. Specifically, this study focuses on evaluating the effectiveness of AI-based strategies that integrate algorithm selection, class balancing methods, and data augmentation approaches to enhance performance for classification of student scientific explanations in the context of an NGSS-aligned assessment.

\noindent\textbf{Research Question:} Which data augmentation strategies best address class imbalance in AI-based text classification models based on model accuracy?

Our study aimed to investigate the impact and the efficacy of three data augmentation strategies on ML model performance in the context of automatic scoring of student responses in science education. We finetuned SciBERT as our base scoring model and evaluated its performance on unbalanced datasets.

\section{Description of Datasets}
The data for this study comes from a previously collected dataset which consists of 1,466 high school student-written responses to a physics three dimensional assessment item designed to elicit ideas about energy \& forces and cause \& effect while constructing an explanation [27]. The item measures student learning on a 3D Learning Progression (LP) with levels 0-3 (Table 1) and focuses on ideas about forces and energy used to construct a causal explanation [29]. The item contains a diagram to show two wood cars with metal sheets attached, with both metal sheets negatively charged. The wedges prevent cars from moving (Figure 1).

\begin{figure}[H]
\centering
\includegraphics[width=0.42\textwidth]{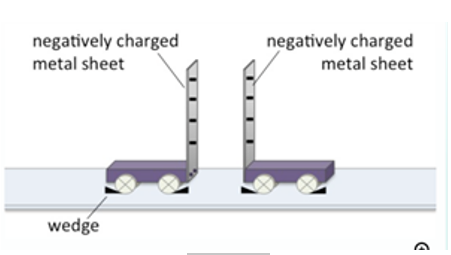}
\caption{Cart item.}
\end{figure}

\noindent Question: When the wedges are removed, the cars will move. Predict which direction they will move and when they will stop. Justify your prediction.

\begin{table}[H]
\centering
\renewcommand{\arraystretch}{1.15}
\begin{tabularx}{\linewidth}{C{0.10\linewidth} Y}
\toprule
Level & Brief Description \\
\midrule
3 & Student models and explanations represent causal relationships that integrate ideas of energy and Coulombic interactions at the atomic-molecular level to explain phenomena. \\
2 & Student models and explanations represent causal relationship that use but do not integrate (or inaccurately integrate) the ideas of energy and/or Coulombic interactions at the macro or atomic-molecular level to explain phenomena with some inaccuracies \\
1 & Student models and explanations represent partially causal relationship that use ideas of Coulombic interactions or energy with inaccurate/incomplete ideas to explain phenomena \\
0 & Student models and explanations don't represent causal relationships and use ideas of Coulombic interactions and/or energy with significantly inaccurate/incomplete ideas \\
\bottomrule
\end{tabularx}
\caption{3D Learning Progression for electrical interactions. Adapted from [27, 29].}
\label{tab:lp}
\end{table}

Student responses were scored using a detailed analytic rubric comprising 11 binary-coded (absence as 0 and presence as 1) categories, with each category targeting either a critical idea from the Learning Progression, or a commonly occurring student misunderstanding related to these concepts [6]. Specifically, categories 1-6 captured the scientific ideas needed to construct an explanation aligned with the top level of the LP and categories 7-11 target inaccurate ideas present in student responses. While we examine classification performance across all 11 categories, at times we focus specific methodologies on categories 5, 6, 7 and 9 as examples within the study. Categories 6 represents integrated understanding of electrical forces and Categories 5 represents energy as described in higher levels of the learning progression. Category 5 captures students' ability to construct a causal relationship using energy concepts only to explain when the carts will stop. Category 6 requires an integrated explanation that links energy concepts with Coulombic interactions to develop a causal explanation. They are the most imbalanced in the dataset because (1) students struggle with core ideas about energy (Category 5), and (2) students struggle to integrate energy concepts with Coulomb's law (Category 6). Category 7 captures responses where a student incorrectly labels an electrostatic (Coulombic) force or field as "magnetic," such as referring to a "magnetic force/field/charge," which indicates confusion between magnetism and electrostatics. Category 9 captures responses that show an incorrect interpretation of Coulomb's law specifically, when students explicitly claim that like charges attract (e.g., "two negatives attract," "they will come together because both are negative") or state an incorrect relationship between force and distance (e.g., saying the force increases as distance increases).

Prior to applying any augmentation strategy, Category 5 contained 1,409 responses labeled as (0) and 57 (3.9\%) responses labeled as (1), while Category 6 contained 1,447 responses labeled as (0) and only 19 responses labeled as (1), resulting in a ratio of approximately 40.103. In response to the Cars item, very few students incorporate ideas about energy, and even fewer incorporate canonical/accurate ideas about energy, specifically via integration with Coulomb's law. As a result, categories 5 and 6 are very unbalanced and have low human-human and human-machine agreement.

\begin{table}[H]
\centering
\scriptsize
\renewcommand{\arraystretch}{1.15}
\begin{tabularx}{\linewidth}{C{0.10\linewidth} Y C{0.16\linewidth}}
\toprule
Rubric Category & Rubric Description & Class imbalance ratio \\
\midrule
1 & Predicts the direction of cart motion, indicating that the carts move away from each other (e.g., repel, separate, move backward, or specifying the direction of each cart). & 5.817 \\
2 & Uses the fundamental property of electric charge to justify motion, explicitly stating that like (negative) charges repel, causing the carts to move apart. & 1.090 \\
3 & Predicts when or whether the carts will stop or slow down, regardless of correctness (e.g., "eventually slow down," "stop when..."). & 1.287 \\
4 & Uses Coulombic reasoning (charge, force, or field) to explain when the carts stop, emphasizing the role of distance in weakening interactions or forces. & 1.725 \\
5 & Explains when the carts stop using energy ideas only, such as energy conversion (potential , kinetic, surroundings) or systems moving toward a minimum energy state, without integrating Coulomb's law. & 19.551 \\
6 & Integrates energy and Coulomb's law to explain motion and stopping, linking high potential energy and strong repulsive forces when close, and decreasing force/energy as distance increases. & 40.103 \\
7 & Incorrectly labels the interaction as magnetic (e.g., magnetic force/field), conflating magnetic and electrostatic phenomena. & 19.818 \\
8 & Shows vague or incorrect use of charge, force, or field terminology, including conflation of these terms (e.g., "charges are weak" instead of force). & 12.13 \\
9 & Demonstrates an incorrect interpretation of Coulomb's law, such as stating that like charges attract or that force increases with distance. & 26.169 \\
10 & Uses energy inaccurately or unproductively, including vague statements ("energy runs out," "energy pushes"), or conflates energy with charge, force, or field. & 3.855 \\
11 & Correctly notes that potential energy is high when carts are close (or decreases as they separate) but fails to explain why (i.e., does not connect to repulsive force or Coulombic reasoning). & 8.957 \\
\bottomrule
\end{tabularx}
\caption{Analytic rubric for the explanation component of the CART items and their corresponding levels of class imbalance.}
\label{tab:rubric}
\end{table}

In the next section, we examine how different augmentation strategies affect these class ratios by addressing imbalance in AI-based text classification models, as reflected in model accuracy.

\section{Methods \& Procedure}
Our first experiment evaluates the baseline approach, using the original dataset with no resampling or augmentation. We then investigate a range of data-augmentation strategies and a resampling technique to address class imbalance and improve text classification performance. Resampling methods are implemented as a comparative benchmark against augmentation-based approaches, allowing us to assess the relative effectiveness of each strategy.

\subsection{Resampling}
Resampling techniques are widely used to address class imbalance in supervised learning by modifying the distribution of training data so that minority and majority classes are more evenly represented [17]. These techniques typically include oversampling, which increases the number of minority-class instances by duplicating existing samples or generating synthetic ones, and undersampling, which reduces the majority class by removing instances. Resampling can improve model sensitivity particularly recall to underrepresented classes by increasing the frequency with which minority examples influence model training [19]. However, resampling also presents limitations. Oversampling via duplication may lead to overfitting when minority classes contain few unique examples, causing models to memorize repeated patterns rather than learn generalizable features [17]. Undersampling, while reducing bias toward majority classes, risks discarding informative data and diminishing the model's ability to capture linguistic and conceptual variability [20]. In educational assessment contexts, where datasets are frequently small and minority classes often represent sophisticated reasoning aligned with higher levels of learning progressions, these limitations motivate alternative strategies, such as data augmentation, that increase class diversity without sacrificing authentic responses.

\subsubsection{Synthetic Minority Over-sampling Technique (SMOTE)}
Chawla et al. [17] developed the Synthetic Minority Over-sampling Technique (SMOTE) algorithm using the feature space of data. The SMOTE algorithm starts with a minority sample with k-nearest neighbors. Assuming that (a, b) is the minority sample and (c, d) is one of the k-nearest neighbors, a new sample can be synthesized as,
\[
(e, f) = (a, b) + \operatorname{rand}(0 - 1) * (c - a, d - b).
\]
A synthetic sample is then created by randomly selecting one of the k-nearest neighbors and interpolating between the two instances in feature space. This process is repeated according to the hyperparameter k until the desired balance between minority and majority classes is achieved.

In the context of this study, SMOTE was employed to balance the distribution of student responses across the 11 analytic rubric categories used for scoring scientific explanations. Our dataset exhibited significant class imbalance, particularly in categories that exhibited a class ratio of greater than 10 (see Table 1 for category-wise distribution). For categories with class ratios greater than 10 (e.g., Categories 5, 6, 7, 9), SMOTE was used to generate synthetic samples until the minority class is balanced with the majority class with a class ratio of approximately 1.

This balancing process was performed separately for each category to ensure that the model could effectively learn the characteristics of both minority and majority classes. The synthetic samples were generated in the feature space of the text, ensuring that the augmented dataset retained the structural textual characteristics of the original student responses. The oversampling rate for each category was determined based on the severity of the imbalance, ensuring that no category was disproportionately oversampled, which could introduce noise or overfitting.

The SMOTE oversampling method on the "Cart item"resample a different number of times for each category to achieve the perfect balance ratio.

\subsection{Augmentation strategy for balancing data}
Figure 2 presents a taxonomy of data augmentation methods in the textual domain, organizing approaches according to the space in which augmentation is performed and the level of linguistic granularity [11]. Broadly, augmentation strategies are divided into feature-space and data-space transformations. Feature-space augmentation operates on processed representations of text, such as embeddings, encoder--decoder outputs, or hidden states of neural networks, rather than on raw textual input. In contrast, data-space augmentation modifies the raw text itself and is therefore more tightly coupled to linguistic structure and task context. As shown in the figure, data-space methods can be further categorized by granularity: character-level approaches (e.g., rule-based edits or noise injection), word-level techniques (e.g., synonym substitution, embedding-based replacements), phrase-level transformations (e.g., interpolation or syntactic restructuring), and document-level methods (e.g., translation or generative rewriting). Data-space augmentation is particularly valuable for text classification tasks, as it preserves semantic content while increasing linguistic diversity. However, the effectiveness of each approach depends strongly on the underlying problem and the need to maintain fidelity to domain-specific meaning, especially in science assessment contexts where conceptual accuracy is critical.

\begin{figure}[H]
\centering
\includegraphics[width=0.92\textwidth]{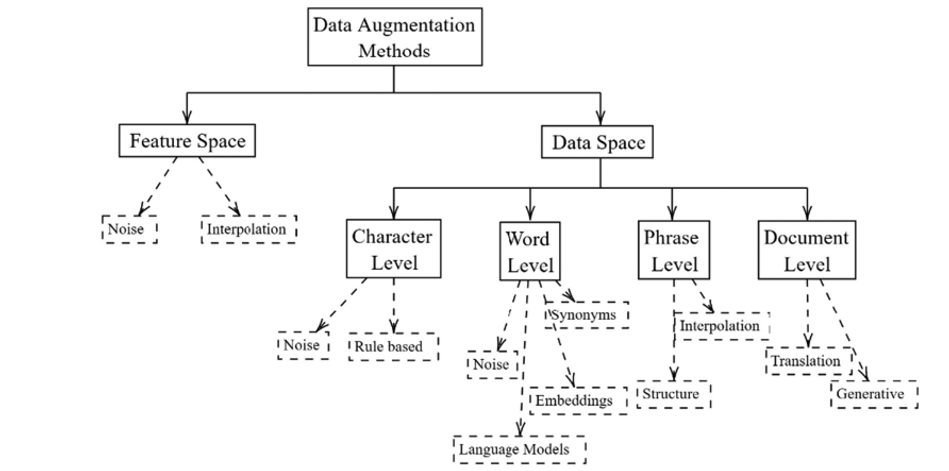}
\caption{Taxonomy and grouping for different data augmentation methods [11].}
\end{figure}

In the scope of this paper and to address the issue of severe class imbalance, we implemented three data-space augmentation techniques, one at the document level (GPT), one at the word level (EASE) and one at the phrase level (ALP), aimed at synthetically enriching the minority class. Examples of responses generated by each augmentation strategy are provided in Appendix 1.

\subsubsection{GPT-Generated Data Augmentation}
This strategy leverages LLMs, specifically GPT-4, to generate synthetic student responses by prompting the model with the original item and examples of answers for a specific characteristic. The generated responses aim to mimic the reasoning and structure of human explanations while introducing slight variations in wording and approach. The goal is to increase the diversity of the training dataset and enhance the model's generalization capability, particularly in low-resource or imbalanced settings [1].

To augment the data, we generated additional response with GPT using 2-Few-Shot Prompting:
\noindent\textbf{2-Few-Shot Prompt:}
\begin{itemize}
\item From the description . Can you understand the following two student's sentence and revise it in your own way?
\item You are a student. Please understand the following sentence and write a similar answer.
\item The following two sentences are students' answers for the above questions separately.
\end{itemize}

"Student N1047: they have the same charge and they will reple with the wedges in place they have potintial energy and once the wedges are removed the for will push them apart then the will have konetic energy"

"Student N1081: If the wedges were to be removed the two cars would move away from each other because the metal sheets are both negatively charged causing them to automatically repel because same charges repel. This happens because the potential energy is built up until it could no longer be stored which is then released and converted to kinetic energy which is why they move away from each other and theyll stop when that energy is transferred to its surroundings causing it to convert thermal energy ."

Now, can you analyze these two students' responses and generate five new suitable answers that are similar responses with the two above's students and satisfy the analytic rubric's criteria for the question?

We record below three GPT-5 generated responses:
\begin{itemize}
\item When the wedges are removed, the carts will start moving because their potential energy is released. That energy turns into kinetic energy, and they will move away from each other. They will stop when all the kinetic energy has been used up or transferred into the environment as heat.
\item Since both carts are negatively charged, they store a lot of potential energy by being close together. When the wedges come out, that energy changes into kinetic energy and makes them move. The motion stops when the system reaches its lowest energy level and there's no more kinetic energy left.
\item The carts will push away from each other after the wedges are taken out because their potential energy becomes kinetic energy. Eventually, that kinetic energy will get smaller as it spreads into the environment, like heat, and the carts will stop when there is no more energy to keep them moving.
\end{itemize}

In our data set, categories 5, 6, 7 and 9 were some of the most severely imbalanced categories. To address the imbalance in Category 5, we applied data augmentation to ensure that the class ratio satisfied a threshold of less than 10:1. Initially, Category 5 contained 1,409 responses labeled as 0 and 57 responses labeled as 1, resulting in a ratio of approximately 24.7:1. To reduce this imbalance, we augmented the minority class until the ratio approached the desired threshold. Specifically, setting the target number of positive instances to 140 yields a ratio of 1,409/140 \$\textbackslash{}approx\$ 10.06. Accordingly, we generated an additional 83 GPT-4 based synthetic responses labeled as 1, resulting in a total of 140 positive instances for Category 5.

To reduce class imbalance, we applied data augmentation to Categories 6, 7, and 9 so that each class ratio fell below the 10:1 threshold. In Category 6, the original ratio was 1,447:19 (approximately 76:1), so we generated 150 GPT-4-based synthetic positive responses, increasing the number of label-1 instances to 169 and reducing the ratio to 1,447:169 (approximately 8.57:1). In Category 7, the original ratio was 1,398:68 (approximately 20.6:1), so we generated 72 synthetic positive responses to reach 140 label-1 instances, yielding a ratio of 1,398:140 (approximately 9.98:1). In Category 9, the original ratio was 1,407:59 (approximately 23.8:1), and we generated additional synthetic positive responses to reach 140 label-1 instances, bringing the ratio to approximately 10:1 (See Table 3).

\begin{table}[H]
\centering
\renewcommand{\arraystretch}{1.15}
\begin{tabularx}{\linewidth}{C{0.18\linewidth} *{4}{C{0.12\linewidth}}}
\toprule
Category & C5 & C6 & C7 & C9 \\
\midrule
Class ratio & 10.06 & 8.57 & 9.98 & 10.06 \\
\bottomrule
\end{tabularx}
\caption{Summary of the level of balance ratio after GPT augmentation.}
\label{tab:gpt-balance}
\end{table}

\subsubsection{EASE (Extract, Acquire, Sift, Employ)}
EASE is a systematic augmentation technique specifically designed to be effective with transformer-based models like SciBERT. Unlike traditional augmentation methods like EDA or AEDA, which often distort syntactic coherence, EASE maintains contextual integrity, making it well-suited for educational text classification where the fidelity of student reasoning is paramount. It follows four key steps [2]:

\begin{description}[style=nextline]
\item[Extract] Meaningful units such as sentences or fact-level statements are extracted from student responses. These fragments preserve the semantic and structural coherence critical for attention-based models.
\item[Acquire] Labels for the extracted fragments are obtained using pretrained models, typically without requiring further human annotation.
\item[Sift] Low-quality or overly short augmented samples are filtered out, ensuring only informative data is retained.
\item[Employ] The curated augmented samples are then merged with the original dataset to expand and enrich the training pool.
\end{description}

Starting from an original dataset of 1,466 responses, EASE generates and filters high-quality synthetic samples for the minority (label = "1'') class in each underrepresented category and then incorporates them into the training set. EASE added 871--1,246 synthetic positive samples per category (e.g., +1,195 for Category 5+1,246 for Category 6; +1,164 for Category 7 ; +1,178 for Category 8 ; +1,187 for Category 9 ; +871 for Category 10 and ; +1,199 for Category 11) to move the distributions toward near parity with the majority class.

We chose to apply augmentation only to Categories 5--11 because Categories 1--4 have class-imbalance ratios in the range of 1--5, indicating they are already relatively well balanced. This motivated us to focus augmentation on Categories 5--11 to improve balance across all categories. Notably, Categories 5, 6, 7, and 9 are the most severely imbalanced within this set; therefore, we also applied other augmentation strategirs to these categories to enable a direct comparison with EASE.

Overall, this process increased the dataset size to 9,504 by adding 8,040 synthetic samples, substantially reducing majority-to-minority imbalance and thereby improving the model's ability to learn minority-class patterns that would otherwise be underrepresented during training.

The following table shows the effect of applying the EASE (Extract--Acquire--Sift--Employ) augmentation pipeline to correct severe class imbalance across several binary label categories from the unbalanced ratios in Table 2.

\begin{table}[H]
\centering
\renewcommand{\arraystretch}{1.15}
\begin{tabularx}{\linewidth}{C{0.18\linewidth} *{7}{C{0.09\linewidth}}}
\toprule
Category & C5 & C6 & C7 & C8 & C9 & C10 & C11 \\
\midrule
Class ratio & 1.13 & 1.14 & 1.13 & 1.05 & 1.13 & 1.08 & 1.08 \\
\bottomrule
\end{tabularx}
\caption{Summary of the level of balance ration after the EASE Augmentation.}
\label{tab:ease-balance}
\end{table}

\subsubsection{Augmentation using Lexicalized Probabilistic (ALP) Context-Free Grammars}
We also used the data Augmentation using Lexicalized Probabilistic (ALP) context-free grammars [10] that generates augmented samples with diverse syntactic structures with plausible grammar paraphrases of each student response in a domain-independent way. Specifically, ALP samples multiple valid plausible subtrees using a probabilistic threshold, extracts plausible subtrees anchored by lexical heads, and produces new sentences by swapping compatible substructures and substituting synonyms (e.g., via WordNet) while preserving the original label/meaning constraints. This approach increases linguistic variability without changing the underlying category label, helping reduce overfitting and improving robustness on imbalanced categories.

Starting from an original dataset of 1,466 responses, ALP generates and filters high-quality synthetic samples for the minority (label = "1'') class in each underrepresented category and then incorporates them into the training set. ALP added 31--405 synthetic positive samples per category (e.g., +95 for Category 5 ; +32 for Category 6; +111 for Category 7 ; +182 for Category 8 ; +98 for Category 9 ; +405 for Category 10 and ; +138 for Category 11) to move the distributions toward near parity with the majority class.

Overall, this process increased the dataset size to 2,525 by adding 1,061 synthetic samples, substantially reducing majority-to-minority imbalance and thereby improving the model's ability to learn minority-class patterns that would otherwise be underrepresented during training. As above we choose to augment only the category from 5 to 11 since the category 1-4 have a balance ratio in the range (1-5).

The following table shows the effect of applying the ALP augmentation pipeline to correct severe class imbalance across several binary label categories from the unbalanced ratio in Table 2.

The primary goal of ALP is to create diversity in the data, allowing the models to train effectively by augmenting the datasets with limited samples. This method is not intended to directly balance the data, which makes it different from EASE.

\begin{table}[H]
\centering
\renewcommand{\arraystretch}{1.15}
\begin{tabularx}{\linewidth}{C{0.18\linewidth} *{7}{C{0.09\linewidth}}}
\toprule
Category & C5 & C6 & C7 & C8 & C9 & C10 & C11 \\
\midrule
Class ratio & 9.33 & 28.33 & 7.80 & 4.60 & 8.95 & 1.84 & 6.32 \\
\bottomrule
\end{tabularx}
\caption{Summary of the level of balance ration after the ALP Augmentation.}
\label{tab:alp-balance}
\end{table}

\subsection{Algorithm Development for Text classification approach}
We implemented a text classification pipeline using SciBERT [4], a domain-specific language model built on the BERT model and pretrained specifically on scientific texts. This is our baseline method. For each analytic rubric category we fine-tuned the SciBERT model [7] for binary classification, aiming to predict whether a given response should receive a score of 0 or 1. Prior research [7] has demonstrated that fine-tuning BERT-based models improves performance on text classification tasks, especially when adapting to domain-specific data split.

The dataset comprises students' textual justifications annotated across multiple categories. Missing values in both the justification text and the categorical label fields were addressed using appropriate imputation procedures. From the 1,466 original student responses, we constructed training and test sets using an 80/20 split. We then trained and evaluated the same model(s) under identical settings and the same train--test split on both the original dataset and the augmented dataset to enable a fair comparison of performance.

Precision, recall, and F1 score are standard metrics for evaluating classification performance. Precision quantifies the proportion of predicted positive instances that are truly positive, while recall (sensitivity) measures the proportion of actual positive instances that the model correctly identifies. The F1 score, defined as the harmonic mean of precision and recall, provides a single summary measure that balances both false positives and false negatives. We use these metrics to assess model performance across the modeling and explanation modalities on the held-out test set.

\section{Analysis \& Findings}
We begin by reporting results for the most severely imbalanced categories. In particular, the first two tables focus on Categories 5, 6, 7, and 9, which exhibit the most extreme skew between the majority and minority classes. In addition, we focus on Categories 5 \& 6 in this rubric as these ideas are critical to identify to assign the highest level of the learning progression. We use these four categories as a stress test to evaluate how well different approaches improve performance, comparing resampling methods against data-augmentation strategies. For these categories, we incorporate both document-level augmentation (GPT-generated synthetic responses) and word- and phrase-level augmentation to assess the relative and combined benefits of these techniques under the most challenging imbalance conditions.

\subsection{Result on extremely imbalanced categories}
\begin{table}[H]
\centering
\scriptsize
\renewcommand{\arraystretch}{1.15}
\begin{tabularx}{\linewidth}{C{0.07\linewidth} C{0.11\linewidth} *{6}{C{0.11\linewidth}}}
\toprule
Category & Metrics & Baseline model: SB & FT SB & SB+ GPT Augmentation & SB+ ALP Augmentation & SB+ EASE Augmentation & SB + SMOTE Over sampling \\
\midrule
5 & Accuracy & 97.61 & 97.26 & 97.70 & 100 & 99.89 & 96.92 \\
5 & Precision & 83.33 & 60.00 & 95.80 & 100 & 99.64 & 56.25 \\
5 & Recall & 45.45 & 81.81 & 79.30 & 100 & 99.88 & 81.81 \\
5 & F1-score & 58.82 & 69.33 & 86.80 & 100 & 99.76 & 66.66 \\
6 & Accuracy & 98.29 & 99.31 & 98.50 & 99.00 & 100 & 98.63 \\
6 & Precision & 0.00 & 100 & 89.70 & 100 & 100 & 50.00 \\
6 & Recall & 0.00 & 33.33 & 92.90 & 60.00 & 100 & 25 \\
6 & F1-score & 0.00 & 50.00 & 91.20 & 75.00 & 100 & 33.33 \\
\bottomrule
\end{tabularx}
\caption{Accuracy, precision, recall, F1 for different strategies on Categories 5 and 6 (Scientific ideas). Note: SB = SciBERT, FT = Fine-tuning.}
\label{tab:cat56}
\end{table}

Table 6 highlights that accuracy remains uniformly high (\$\textbackslash{}approx\$96.9--100) across strategies, but this masks major differences in how well models actually detect the scientific-idea labels (Categories 5 and 6). In Category 6, the baseline SciBERT collapses entirely---despite 98.29 accuracy, it records 0.00 precision, 0.00 recall, and 0.00 F1, indicating it essentially never identifies the minority class. Fine-tuning partially corrects this failure (precision rises to 100), but recall remains low (33.33), producing only F1 = 50.00---a conservative model that is correct when it predicts the class, yet still misses most true scientific ideas. In Category 5, fine-tuning improves recall substantially (45.45 to 81.81) and raises F1 (58.82 to 69.33), but precision drops (83.33 to 60.00), showing that the model becomes more sensitive but less selective.

We observe that all augmentation strategies consistently improve minority-class learning over the baseline and stabilizes the precision--recall balance, yielding much higher F1 across both categories. For Category 6, SB+GPT achieves F1 = 91.20 (precision 89.70, recall 92.90), showing robust recovery from baseline collapse, while SB+ALP is more precision-heavy (100 precision, 60 recall; F1 = 75.00). For Category 5, augmentation also raises performance sharply (e.g., SB+GPT F1 = 86.80), and SB+EASE is near-perfect (F1 = 99.76) with SB+ALP reaching 100 across all metrics in this table. By contrast, SMOTE oversampling is comparatively weaker, especially in Category 6 (F1 = 33.33) and Category 5 (F1 = 66.66), suggesting that embedding-space resampling does not provide the same quality of class-specific signal as language-grounded augmentation. Overall, the table supports the conclusion that targeted text augmentation is more effective than resampling alone for recovering scientific-idea categories under extreme imbalance.

\begin{table}[H]
\centering
\scriptsize
\renewcommand{\arraystretch}{1.15}
\begin{tabularx}{\linewidth}{C{0.07\linewidth} C{0.11\linewidth} *{6}{C{0.11\linewidth}}}
\toprule
Category & Metrics & Baseline model: SB & FT SB & SB+ GPT Augmentation & SB+ EASE Augmentation & SB+ ALP Augmentation & SB + SMOTE Over sampling \\
\midrule
7 & Accuracy & 98.29 & 99.31 & 97.40 & 99.63 & 100 & 99.65 \\
7 & Precision & 92.30 & 100 & 100 & 98.68 & 100 & 93.33 \\
7 & Recall & 75.00 & 87.50 & 75.76 & 99.01 & 100 & 100 \\
7 & F1-score & 82.75 & 93.33 & 86.21 & 98.85 & 100 & 96.55 \\
9 & Accuracy & 92.49 & 93.17 & 95.50 & 99.60 & 99.50 & 95.56 \\
9 & Precision & 0 & 0 & 77.78 & 97.55 & 100 & 40.00 \\
9 & Recall & 0 & 0 & 58.33 & 99.61 & 83.33 & 16.67 \\
9 & F1-score & 0 & 0 & 66.67 & 98.57 & 90.91 & 23.52 \\
\bottomrule
\end{tabularx}
\caption{Accuracy, precision, recall, F1 for different strategies on Categories 7 and 9 (Non Scientific ideas). Note: SB = SciBERT, FT = Fine-tuning.}
\label{tab:cat79}
\end{table}

We then examined whether the set of augmentation strategies performed similarly for severely imbalanced classes related to inaccurate ideas. In Category 7, performance is already strong for the baseline SciBERT (Accuracy = 98.29; Precision = 92.30; Recall = 75.00; F1 = 82.75), and fine-tuning further improves minority-class detection, raising recall to 87.50 and F1 to 93.33 while reaching 100 precision. However, GPT augmentation does not improve Category 7 relative to fine-tuning: accuracy drops to 97.40 and recall remains around the baseline level (75.76), yielding a lower F1 (86.21) than FT. In contrast, the EASE and ALP augmentation techniques produce the strongest and most stable outcomes, with SB+EASE reaching near-ceiling performance (Recall = 99.01; F1 = 98.85) and SB+ALP achieving perfect scores (100) across all metrics. SMOTE also performs well for Category 7 (Recall = 100; F1 = 96.55), suggesting this category is comparatively easier to learn once sufficient minority signal is available.

In Category 9, the table shows a much harsher imbalance-driven failure mode: both the baseline and fine-tuned models show non-informative minority-class performance (Precision = 0, Recall = 0, F1 = 0) despite seemingly reasonable accuracy (92.49--93.17), indicating the models effectively default to the majority class and miss Category 9 entirely. Augmentation fundamentally changes this: GPT augmentation somewhat recovers minority detection (Precision = 77.78; Recall = 58.33; F1 = 66.67), but the largest gains come again from EASE and ALP, with SB+EASE delivering near-perfect results (Accuracy = 99.60; Precision = 97.55; Recall = 99.61; F1 = 98.57) and SB+ALP also very strong (Precision = 100; Recall = 83.33; F1 = 90.91). By comparison, SMOTE oversampling remains relatively weak for Category 9 (Precision = 40.00; Recall = 16.67; F1 = 23.52), reinforcing the finding that language-grounded augmentation (especially EASE and/or ALP) is most effective for rescuing extreme minority classes, while oversampling alone does not reliably prevent collapse.

Furthermore, we also fine-tuned SciBERT models using the GPT-based augmentation approach for Categories 5, 6, 7, and 9. We then report accuracy, precision, recall, and F1-score to evaluate performance (see Appendix 2). Overall, the fine-tuned models show improved accuracy compared with the baseline.

\subsection{Results across all categories}
Table 8 reports the average performance across all categories and summarizes overall agreement between the automated scoring models and human annotations. For interpretability, we group the categories into two broad sets: scientific ideas (codes reflecting scientifically accurate reasoning) and inaccurate ideas (codes capturing misconceptions or incorrect reasoning). This organization allows us to examine whether the model more reliably identifies correct scientific reasoning or common inaccuracies, and to determine which set yields stronger alignment with human scoring.

\begin{table}[H]
\centering
\scriptsize
\renewcommand{\arraystretch}{1.15}
\begin{tabularx}{\linewidth}{C{0.10\linewidth} C{0.11\linewidth} *{5}{C{0.12\linewidth}}}
\toprule
Category & Metrics & Baseline model: SciBERT & FT SB & SB+ ALP & SB+ EASE Augmentation & SB + SMOTE \\
\midrule
1-6 & Accuracy & 92.03 & 96.13 & 99.08 & 99.47 & 95.96 \\
1-6 & Precision & 72.22 & 88.91 & 98.72 & 99.36 & 80.03 \\
1-6 & Recall & 67.20 & 83.30 & 92.89 & 99.45 & 81.96 \\
1-6 & F1-score & 68.75 & 83.02 & 94.96 & 99.41 & 79.84 \\
7-11 & Accuracy & 93.31 & 95.22 & 99.40 & 99.63 & 95.56 \\
7-11 & Precision & 47.69 & 58.96 & 98.34 & 98.76 & 64.64 \\
7-11 & Recall & 36.24 & 57.17 & 92.16 & 99.33 & 62.17 \\
7-11 & F1-score & 40.99 & 57.87 & 95.00 & 99.04 & 62.28 \\
\bottomrule
\end{tabularx}
\caption{Accuracy, precision, recall, F1 for different strategies for all Categories (divided by Scientific ideas and non Scientific ideas). Note: SB = SciBERT, FT = Fine-tuning.}
\label{tab:allcats}
\end{table}

Across Categories 1--6 (Scientific ideas), Table 6 shows a clear progression from baseline to augmentation-driven approaches. The baseline SciBERT already attains high accuracy (92.03), but its F1-score (68.75) indicates weaker minority-class identification than accuracy suggests. Fine-tuning improves all metrics (Precision 88.91, Recall 83.30, F1 83.02), confirming that task-adaptive training helps the model better capture scientific-idea patterns. However, the largest gains come from the phrase and word augmentation strategies: SB+ALP pushes performance to near-ceiling levels (Precision 98.72, Recall 92.89, F1 94.96), while SB+EASE delivers the strongest overall results (Accuracy 99.47, Precision 99.36, Recall 99.45, F1 99.41), indicating highly robust and balanced detection of scientific ideas. SMOTE, in contrast, improves over baseline but remains clearly behind the augmentation strategies across all metrics.

For Categories 7--11 (inaccurate ideas), the same pattern is even more pronounced, showing that these categories are harder for the baseline model to detect reliably. Although baseline accuracy is high (93.31), minority-class performance is poor (Precision 47.69, Recall 36.24, F1 40.99), showing that the model struggles to capture non-scientific ideas despite appearing accurate overall. Fine-tuning helps (F1 57.87), but augmentation transforms performance: SB+ALP reaches F1 = 95.00 (Precision 98.34, Recall 92.16), and SB+EASE again achieves near-perfect outcomes (Accuracy 99.63, Precision 98.76, Recall 99.33, F1 99.04). Notably, SMOTE performs modestly (F1 62.28) but is far behind improvements via ALP and EASE approaches. This suggests that augmentation strategies provide consistent and substantial improvements across both scientific and non-scientific groups, while resampling methods offer limited benefit under severe imbalance. Overall, this experiment demonstrates that a targeted augmentation strategy can robustly improve classification performance across a range of rubrics, and imbalance levels.

\section{Conclusion and Implications}
This work contributes to advancing automated scoring of student explanations including complex student reasoning using AI tools. Our study targets a pressing need in science assessment, the automated classification of open-ended responses using analytic rubrics that reflect learning progressions, which reflect various sophistication of student understanding while learning. We examined which augmentation strategies most effectively mitigate class imbalance in transformer-based automated scoring of student scientific explanations aligned with a learning progression (LP). Using SciBERT as the base classifier, we compared a no-augmentation baseline and fine-tuned model against resampling (SMOTE) and three data-space augmentation approaches (GPT-generated document-level augmentation, EASE word-level augmentation, and ALP phrase-level augmentation).

Our results show that targeted data augmentation is the most reliable solution for severe imbalance, especially in the most conceptually advanced and rare categories (e.g., Categories 5--6, which represent higher-level energy and Coulombic reasoning and are among the most imbalanced and instructionally critical for LP placement). In the extreme-imbalance setting, fine-tuning alone improved recall but still left important gaps, whereas GPT augmentation improved both precision and recall, and ALP delivered the strongest performance, achieving perfect precision/recall/F1 in the most skewed categories (Categories 5, 6,7 and 9).

By focusing on imbalanced but conceptually rich categories like reasoning about energy and Coulombic interactions (Categories 5 and 6), we respond to a common issue in classroom data: small datasets skewed toward novice reasoning. These data limitations often undermine efforts to validly assess deeper levels of scientific understanding, including the use of LP-aligned assessments. Our integration of SciBERT with augmentation strategies (specifically, ALP and EASE) yielded high classification performance even under conditions of severe class imbalance. These results represent a methodological advance for educators and researchers seeking scalable approaches to using three dimensional assessments aligned to LPs. Importantly, these categories play a central role for assigning students to advanced levels of the LP, underscoring the importance of accurate scoring of these categories. Accurate assignment to LP levels is not only about scoring precision but ensuring that students are correctly placed in the LP in ways that advance the next steps for instruction and learning. In this way, improved classification accuracy supports more targeted feedback and more relevant opportunities for students to advance their learning based on their current understanding.

Notably, the same augmentation strategies also improved predictive performance for Categories 7 and 9. Although these categories also occur infrequently, they capture consequential inaccuracies in student reasoning: one reflects conflation of magnetic and electric forces, while the other indicates a misunderstanding of Coulombic interactions in the item context. Because these inaccurate ideas are lexically and linguistically similar to parts of normative scientific explanations, they are particularly difficult to detect using automated scoring approaches. Yet identifying these ideas is essential, as advancing along the LP requires that such inaccurate ideas be explicitly addressed through student revision and instructional support. The observed improvements in model performance with ALP and EASE therefore suggest that these augmentation strategies may be broadly applicable for identifying a wide range of targeted, conceptually nuanced ideas, both normative and non-normative, in student explanations.

This is further supported by examination of the average performance of different strategies across all categories, exhibiting a range of balance. Across all rubric categories, augmentation strategies, particularly EASE, substantially increased alignment with human scoring for both scientific ideas (Categories 1--6) and inaccurate ideas (Categories 7--11), yielding near maximal performance relative to the baseline and fine-tuning alone. ALP augmentation exhibited similar performance as EASE on both scientific and inaccurate ideas, however exhibited lower recall. This is critical since it means ALP is missing more positive cases of some of these inaccurate ideas, so responses that display such inaccuracies may not be properly labeled. Nevertheless, the EASE and ALP overall performance was strong. Moreover, because these augmentation strategies can be implemented automatically within a text classification pipeline, they reduce reliance on additional data collection or manual labeling of student responses.

A key implication is that resampling alone is not sufficient for LP-aligned automated scoring under extreme imbalance. While resampling can increase minority-class exposure, it can also introduce artifacts: oversampling may encourage memorization when few unique minority examples exist, and undersampling can remove valuable novice-level responses needed to preserve conceptual coverage along the LP. Consistent with this limitation, SMOTE underperformed relative to augmentation in our aggregate results, plausibly due to overfitting and reduced linguistic/semantic diversity when balancing is achieved primarily through feature-space interpolation rather than generating meaning-preserving variations in students' natural language.

For practice and future development, these findings suggest that augmentation-centered pipelines utilizing EASE and ALP can make LP-aligned scoring more scalable and instructionally useful. They improve detection of rare but critical forms of sophisticated reasoning (Categories 5--6) and also strengthen identification of subtle misconceptions (e.g., misinterpreting Coulomb's law), which are essential targets for formative feedback and revision. At the same time, the use of synthetic data requires additional validity work. Future studies should test generalization across additional items, rubrics, domains, and student populations and incorporate systematic quality checks by experts to ensure that augmented responses preserve the intended construct.

\section{Code avaibility}
Code, trained adapters, and detailed training logs are publicly available. All experiments are publicly available [9]:
\url{https://github.com/Prud11djagba/-Optimizing-AI-Scoring-of-Scientific-Explanations-Exploring-Augmentation-Strategies-}

\section{Acknowledgements}
This material is based upon work supported by the National Science Foundation under Grant No. 2200757. Any opinions, findings, and conclusions or recommendations expressed in this material are those of the author(s) and do not necessarily reflect the views of the National Science Foundation.

\section{Appendix 1: Actual examples of augmented responses}
\scriptsize
\renewcommand{\arraystretch}{1.15}
\begin{longtable}{C{0.07\linewidth} p{0.88\linewidth}}
\caption{Examples of ALP generated responses.}\\
\toprule
N & Generated responses with ALP \\
\midrule
\endfirsthead
\caption[]{Examples of ALP generated responses. (continued)}\\
\toprule
N & Generated responses with ALP \\
\midrule
\endhead
\bottomrule
\endfoot
1 & They are unstable so they will gained and they will increases when they are far apart because so that means they will repel from each other . \\
2 & I suppose that since they are the high ( they are both somewhere charged ) they with move the same way and wo n't stop until they get to no forces between them . \\
3 & When the aka are known the object would move away from each other since they have the same charges . Being both negative are both a negative charge and like charges repel . The would both stop when the charge ca n't reach each other . \\
4 & When the whole will given less from each other because they both have negatively charged plates on them and same charges repel , they will move away from each other because they both have negatively charged plates on them and same charges repel . They move that direction , because both the metal sheets are negatively charged , We know that like charges repel , so depending on how strong the negative charge is , is how far the carts will move away from each other . \\
5 & They will comes and attracted the remove way from a strong force between them . Considering that they are opposites they will not connect whatsoever . They would most likely stop when they will be a strong force between them two probably making the carts stop when they are really far away from each other . \\
6 & The simulation attracting the repel.they because the cars have the same charge negative charge . They will push away until there is no more energy then they are negatively charged \\
7 & They will be balanced twice from each opposite , and both negative and same signs repeal will stop when thermal energy of the energy is too weak to continue pushing them away from each other . \\
8 & The pulls attracts would cause away from each other until the charges had The cars anymore , this being , it would split in directions . \\
9 & When the kine are plated the strength will stop moving when the force between them becomes too weak and they will stop when they leave their electric fields . \\
10 & When you stick the position from the negative they will both move in a opposite direction because they are both the same charge as each other . The force of the charge is great enough do n't effect each other anymore . Once they start moving they are now kinetic energy . \\
\end{longtable}

\scriptsize
\renewcommand{\arraystretch}{1.15}
\begin{longtable}{C{0.07\linewidth} p{0.88\linewidth}}
\caption{Examples of EASE generated responses.}\\
\toprule
N & Generated responses with EASE \\
\midrule
\endfirsthead
\caption[]{Examples of EASE generated responses. (continued)}\\
\toprule
N & Generated responses with EASE \\
\midrule
\endhead
\bottomrule
\endfoot
1 & They move that direction, because both the metal sheets are negatively charged, We know that like charges repel, so depending on how strong the negative charge is, is how far the carts will move away from each other. \\
2 & when the wedges move the carts would move away from each other because like charges do not attract. \\
3 & I think that the cars will move away from each other due to both of the sheets having the same charge of the other, and will only come to a stop upon hitting a hall, loosing momentum, or coming back around to face the opposite vehicle sheet to sheet. \\
4 & They will stop when they leave the magnetic field around them the magnetic field is very small when they are close together but the farther away they get from each other the weaker the field will get. \\
5 & The cars will move away from each other cause they are both negative and so opposites repel so the forces and energy will push them away from each other. \\
6 & The two carts will repel from one another by different forces and end up the same distance apart because the energy and force between the same charged plates will make the carts have the same repelling force. \\
7 & what think will happen is that when the wedges are removed the two car will push away from each other because the have the same charge and when u have the same charge they repel \\
8 & The cars will move opposite direction from one another because the forces are negative in both cars and the same charges repel , there isn't a time that could be predicted but once the energy runs out they will slowly stop running and once there isn't a force pulling it or repelling it , it won't have a force that will make it move . \\
9 & because the plates are both negative charged, when the wedges are removed the cars will start moving away from each other in opposite directions and wont stop until the negative forces can no longer act upon each other when they are far enough away from each other \\
10 & once the wedges get removed the cars will push away from each other because they are both negatively charged and will repel. \\
\end{longtable}

\scriptsize
\renewcommand{\arraystretch}{1.15}
\begin{longtable}{C{0.07\linewidth} p{0.88\linewidth}}
\caption{Examples of GPT generated responses.}\\
\toprule
N & Generated responses with GPT-5 \\
\midrule
\endfirsthead
\caption[]{Examples of GPT generated responses. (continued)}\\
\toprule
N & Generated responses with GPT-5 \\
\midrule
\endhead
\bottomrule
\endfoot
1 & The cars move because of repulsion from similar charges. The energy is high when they're close and goes down as they move farther apart until there is no field pushing them anymore. They stop when the energy is at its lowest. They continue to move until the field is minimized. \\
2 & since the cars are negatively charged, the potential energy between them will turn into kinetic energy as they move apart. but once the energy is gone, they'll just stop cause they don't have anymore force pushing them. \\
3 & the cars start moving cause the energy between them is high, and it turns into kinetic energy as they move away. as they get farther, the system's energy goes down but its not clear if they'll ever actually stop. \\
4 & they move apart because they're trying to reach a low energy state. the potential energy is high when they're close and decreases as they get farther apart. they stop when the field is at its lowest. they will keep moving until the energy is gone, reaching a stable position. \\
5 & the repulsive force between the carts is what makes them move when the wedges are removed. when they're close, they have a lot of energy, but as the carts move apart, both the energy and force decrease until they can't move anymore. \\
6 & when the wedges come out, the carts will start moving because their potential energy is really high. the kinetic energy increases as they move apart, but they won't stop until the energy between them is used up. \\
7 & they push each other away because of the repulsion field between them. both carts are negatively charged, and when close, they have high energy. once they start moving apart, the energy decreases till they stop moving. they will stop when the field is gone and no repulsive force remains. \\
8 & The carts have potential energy stored because they're being held close while repelling. Once the wedges are taken out, that energy becomes kinetic energy and makes them move. Over time, the kinetic energy is lost as heat, and the carts eventually stop. \\
9 & At first, the carts have high potential energy because they are close together and repelling. When released, this energy becomes kinetic energy. They stop moving when all the kinetic energy is transferred to the surroundings. \\
10 & The carts are forced close together while having the same charge, which creates potential energy. When the wedges are taken out, that energy turns into kinetic energy, making them move. They stop when the energy is fully transferred into the environment as heat. \\
\end{longtable}

\section{Appendix 2}
\scriptsize
\renewcommand{\arraystretch}{1.15}
\begin{longtable}{C{0.08\linewidth} C{0.11\linewidth} *{4}{C{0.14\linewidth}}}
\caption{Accuracy, precision, recall, and F1-score for Categories 5, 6, 7, and 9 across the different strategies, using SciBERT fine-tuned with the GPT-based augmentation approach. Note: SB = SciBERT, FT = Fine-tuning.}\\
\toprule
Category & Metrics & Baseline model: SB & FT SB & SB+ GPT Augmentation & FT SB + GPT Augmentation \\
\midrule
\endfirsthead
\caption[]{Accuracy, precision, recall, and F1-score for Categories 5, 6, 7, and 9 across the different strategies, using SciBERT fine-tuned with the GPT-based augmentation approach. Note: SB = SciBERT, FT = Fine-tuning. (continued)}\\
\toprule
Category & Metrics & Baseline model: SB & FT SB & SB+ GPT Augmentation & FT SB + GPT Augmentation \\
\midrule
\endhead
\bottomrule
\endfoot
5 & Accuracy & 97.61 & 97.26 & 97.70 & 97.75 \\
5 & Precision & 83.33 & 60.00 & 95.80 & 86.67 \\
5 & Recall & 45.45 & 81.81 & 79.30 & 89.66 \\
5 & F1-score & 58.82 & 69.33 & 86.80 & 88.14 \\
6 & Accuracy & 98.29 & 99.31 & 98.50 & 99.06 \\
6 & Precision & 0.00 & 100 & 89.70 & 93.10 \\
6 & Recall & 0.00 & 33.33 & 92.90 & 96.43 \\
6 & F1-score & 0.00 & 50.00 & 91.20 & 94.74 \\
7 & Accuracy & 98.29 & 99.31 & 97.40 & 99.35 \\
7 & Precision & 92.30 & 100 & 100 & 94.29 \\
7 & Recall & 75.00 & 87.50 & 75.76 & 100 \\
7 & F1-score & 82.75 & 93.33 & 86.21 & 97.06 \\
9 & Accuracy & 92.49 & 93.17 & 95.50 & 97.75 \\
9 & Precision & 0 & 0 & 77.78 & 90.48 \\
9 & Recall & 0 & 0 & 58.33 & 79.17 \\
9 & F1-score & 0 & 0 & 66.67 & 84.44 \\
\end{longtable}

Table 9 reports the accuracy, precision, recall, and F1-score for Categories 5, 6, 7, and 9 across the different strategies, using SciBERT fine-tuned with the GPT-based augmentation approach. Overall, the fine-tuned models demonstrate improved accuracy compared with the baseline.


\end{document}